\pdfoutput=1

\documentclass[11pt]{article}

\usepackage[final]{acl}
\usepackage{verbatim} 
\usepackage{hyperref}

\usepackage{lscape}
\usepackage{tabularx}
\usepackage{longtable}
\usepackage{times}
\usepackage{latexsym}

\usepackage[T1]{fontenc}

\usepackage[utf8]{inputenc}

\usepackage{microtype}

\usepackage{inconsolata}
\usepackage{amsmath}
\usepackage{amssymb}
\usepackage{cases}
\usepackage{graphicx}

\title{Medifact at PerAnsSumm 2025: Leveraging Lightweight Models for Perspective-Specific Summarization of Clinical Q\&A Forums}

\author{Nadia Saeed \\
  Computational Biology Research Lab \\ Department of Computer Science \\
  National University of
Computer and Emerging Sciences (NUCES-FAST)\\ Islamabad, Pakistan \\
  \texttt{i181606@nu.edu.pk} \\}

\begin{document}
\maketitle
\begin{abstract}
The PerAnsSumm 2025 challenge focuses on perspective-aware healthcare answer summarization \cite{peransumm-overview}. This work proposes a few-shot learning framework using a Snorkel-BART-SVM pipeline for classifying and summarizing open-ended healthcare community question-answering (CQA). An SVM model is trained with weak supervision via Snorkel, enhancing zero-shot learning. Extractive classification identifies perspective-relevant sentences, which are then summarized using a pretrained BART-CNN model. The approach achieved 12th place among 100 teams in the shared task, demonstrating computational efficiency and contextual accuracy. By leveraging pretrained summarization models, this work advances medical CQA research and contributes to clinical decision support systems.\footnote{Models Code available: \url{https://github.com/NadiaSaeed/PerAnsSumm2025/tree/main}}
\end{abstract}

\section{Introduction}

Healthcare Community Question-Answering (CQA) forums have become a vital source of medical information to seek advice and share experiences \cite{jiang2024clinical,zhang2024development}. These platforms generate diverse responses, ranging from factual knowledge to personal opinions like PUMA dataset \cite{naik-etal-2024-perspective}. Traditional CQA summarization methods focus on selecting a single best-voted answer as a reference summary \cite{tsatsaronis2015overview,kell2024question}. However, a single answer often fails to capture the broad range of perspectives available across multiple responses. To better serve users, it is essential to generate structured summaries that represent various viewpoints effectively.

To address this, we introduce a hybrid framework that combines perspective classification and summarization, as shown in Figure \ref{fig:02}. The first step involves classifying user responses into predefined perspectives using a multi-step learning pipeline. This pipeline integrates Snorkel-based weak supervision \cite{ratner2017snorkel}, support vector machine (SVM) classification with sentence embeddings \cite{rueping2010svm}, and zero-shot learning (ZSL) using transformer models \cite{lewis2019bart}. The goal is to enhance classification accuracy, especially when labeled data is scarce.

Once classified, responses undergo a two-step summarization process. We employ extractive summarization using BART to select key sentences from classified perspectives \cite{lewis2019bart}. Then, we refine these summaries using abstractive summarization with Pegasus to improve fluency and coherence~\cite{zhang2020pegasus}. The composed model is evaluated on the \textbf{PerAnsSumm Shared Task - CL4Health@NAACL 2025}, which focuses on analyzing multi-perspective responses in Community Question Answering (CQA) \cite{peransumm-overview}. Given a user-generated question \( Q \) and a set of responses \( A \), the task is divided into two key objectives: 

(1)Perspective Classification, where response spans are categorized into predefined perspectives such as \textit{cause}, \textit{suggestion}, \textit{experience}, \textit{question}, and \textit{information}; 

(2)Perspective Summarization, which generates structured summaries that condense key insights while preserving essential details. Our approach integrates both tasks into a single pipeline, ensuring efficient classification and summarization of CQA responses. 

By leveraging weak supervision and fine-tuning pre-trained models, we balance computational efficiency with adaptability, making the solution practical for real-world applications.  
 This hybrid approach ensures that summaries retain critical information while being concise and easily understandable. This study makes the following key contributions:
 \begin{figure*}[h]
  \includegraphics[width=\textwidth]{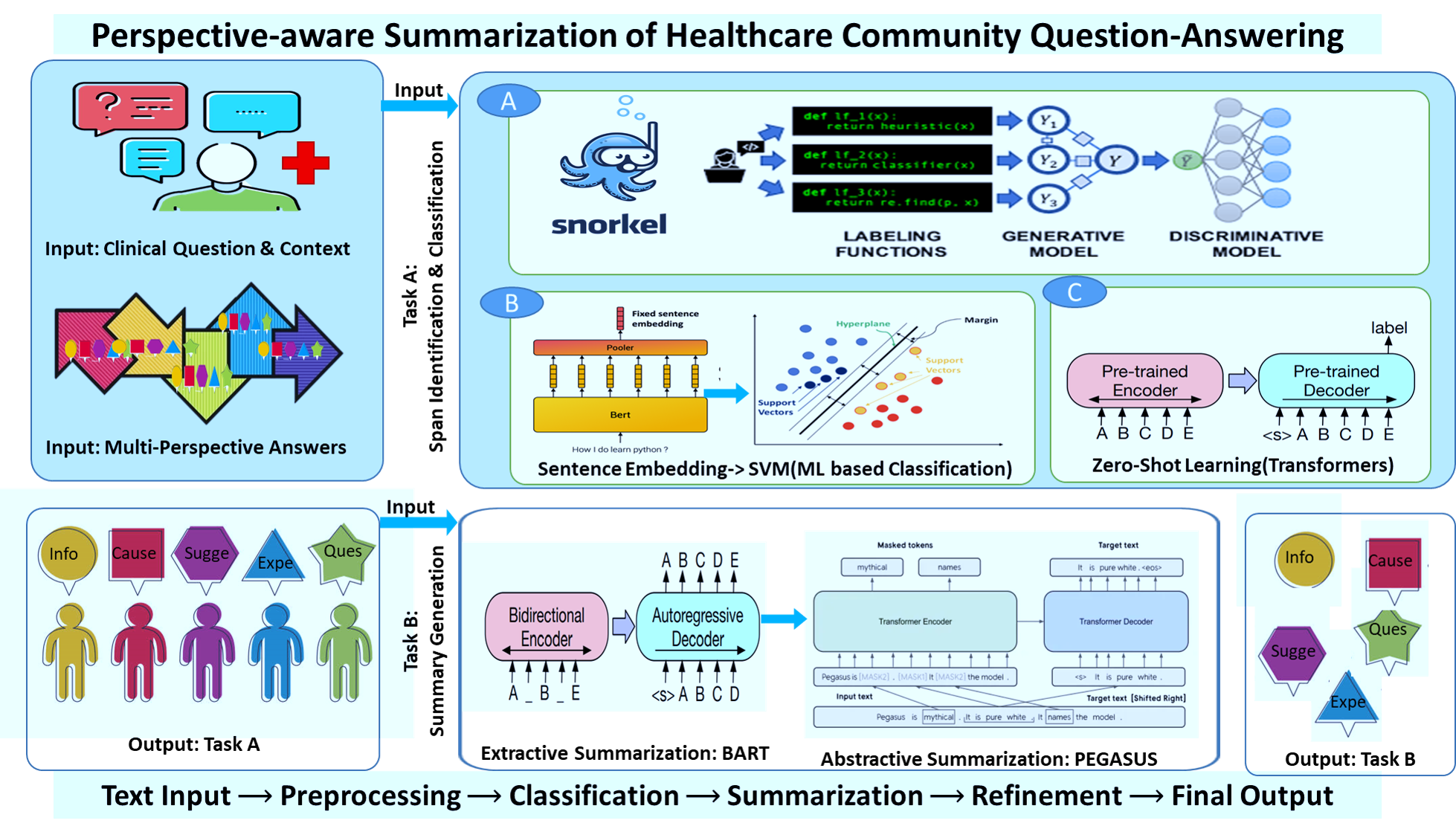}
  \caption{Hybrid workflow for perspective classification and summarization. Perspectives are classified using heuristic labeling (Snorkel), SVM-based classification, and a zero-shot model fallback. Summarization is performed in two stages: extractive (BART) and abstractive (Pegasus), integrating the context for a refined output.}
  \label{fig:02}
\end{figure*}
\begin{enumerate}
\item A hybrid classification framework combining weak supervision, machine learning, and deep learning techniques for robust perspective identification.
\item A rule-based weak supervision method using Snorkel’s labeling functions to generate high-quality probabilistic labels.
\item Feature extraction via sentence embeddings, leveraging transformer-based models to enhance classification.
\item A zero-shot learning (ZSL) classifier to handle unseen data without additional labeled examples.
\item A two-stage summarization pipeline that integrates extractive (BART) and abstractive (Pegasus) techniques for structured summaries.
\item A thorough evaluation demonstrating the effectiveness of our approach on real-world CQA datasets.
\end{enumerate}

By combining classification with summarization, our method ensures that user-generated responses are structured, informative, and accessible. This enhances the usability of healthcare CQA forums and facilitates better decision-making for users.

\section{Methodology}

\subsection{Task A: Perspective Classification}

\subsubsection{Problem Definition}
Given a dataset of textual responses, our goal is to classify each response $x_i$ into one of the predefined perspective categories \cite{naik-etal-2024-perspective}:
\begin{equation}
\mathcal{P} = \{ \textit{EXPE},\textit{INFO}, \textit{CAUS},\textit{SUGG},\textit{QUES} \}
\label{eq.1}
\end{equation}

Each response consists of multiple sentences, and our objective is to determine the category $y_i$ by maximizing the conditional probability:

\begin{equation}
 y_i = \arg\max_{p \in \mathcal{P}} P(p \mid x_i)
\end{equation}

\subsubsection{Hybrid Classification Pipeline}
To achieve robust classification, we employ a three-stage hybrid pipeline:

\begin{enumerate}
    \item Weak Supervision with Snorkel: Rule-based labeling functions assign probabilistic labels \cite{ratner2017snorkel,fries2020trove,ruhling2021end}.
    \item Supervised Learning with SVM: A Support Vector Machine (SVM) refines classification using sentence embeddings \cite{ala2023multilingual}.
    \item Zero-Shot Classification: A transformer model is applied when previous methods yield uncertain labels \cite{gera2022zero,schopf2022evaluating}.
\end{enumerate}

\subsubsection{Weak Supervision Using Snorkel}
Manual annotation is time-intensive, so we use Snorkel’s labeling functions (LFs) to generate weak labels based on pattern recognition:

\begin{equation}
 LF(x) = \begin{cases} l_p, & \text{if pattern } p \text{ is found in } x \\ -1, & \text{otherwise} \end{cases}
\end{equation}

where $l_p$ is the assigned label, and $-1$ indicates abstention. To aggregate multiple weak labels, Snorkel’s Label Model $M$ estimates the true label distribution:

\begin{equation}
 \hat{Y} = M(L)
\end{equation}

where $L$ represents the label matrix from different LFs. To efficiently label textual data, LFs based on regex patterns extracted from frequent words in the dataset. Each LF detects specific linguistic cues for perspective categories like EXPERIENCE or SUGGESTION. If a match is found, a label is assigned; otherwise, it abstains (as shown in Figure \ref{fig:02}). The PandasLFApplier applies these LFs to generate a label matrix \cite{tok2021practical}, which is then refined using Snorkel’s Label Model to resolve conflicts and improve accuracy. This approach speeds up annotation while ensuring consistency through statistical aggregation.

\subsubsection{Sentence Embeddings and SVM Classification}
We convert textual responses into sentence embed
\begin{equation}
 E(x) = \text{SentenceTransformer}(x)
\end{equation}
These embeddings are used by an SVM classifier to enhance prediction accuracy:
\begin{equation}
 \hat{y} = \text{SVM}(E(x))
\end{equation}
SVM is trained on sentence embeddings from a labeled dataset to classify text into perspective categories. Using a linear kernel, it learns decision boundaries in high-dimensional space. During inference, new sentences are embedded and classified based on their positions in the learned feature space.
\subsubsection{Few-Shot Learning with Zero-Shot Classification}
If Snorkel and SVM fail to provide a confident classification, we apply zero-shot learning (ZSL) using a transformer-based model:
\begin{equation}
 P(p \mid x) = f_{ZSL}(x, \mathcal{P})
\end{equation}
where $f_{ZSL}$ is a BART-based ZSL classifier, selecting the category with the highest probability. The ZSL model (facebook/bart-large-mnli) is applied using Hugging Face’s pipeline \cite{lewis2019bart}. When a sentence remains unclassified, the ZSL model evaluates the text without prior training on specific labeled data by comparing it to predefined perspective categories (\textit{P}). It then assigns the most probable label by ranking all categories based on their semantic similarity to the input sentence. This ensures that even unseen or ambiguous responses can still be categorized effectively. 

\subsubsection{Final Classification Decision}
The classification decision follows a hierarchical approach (as Shown in Figure \ref{fig:02} A, B and C):

\begin{equation}
 y_i = \begin{cases}
 \hat{Y}_i, & \text{if } \hat{Y}_i \neq -1 \\
 \text{SVM}(E(x_i)), & \text{if Snorkel abstains} \\
 f_{ZSL}(x_i, \mathcal{P}), & \text{otherwise}
 \end{cases}
 \label{eq.8}
\end{equation}

\subsection{Task B: Hybrid Summarization}

\subsubsection{Overview}
To generate high-quality summaries, we integrate extractive and abstractive techniques as shown in Figure \ref{fig:02} and \ref{fig:03}:
\begin{figure*}[h]
  \includegraphics[width=\textwidth]{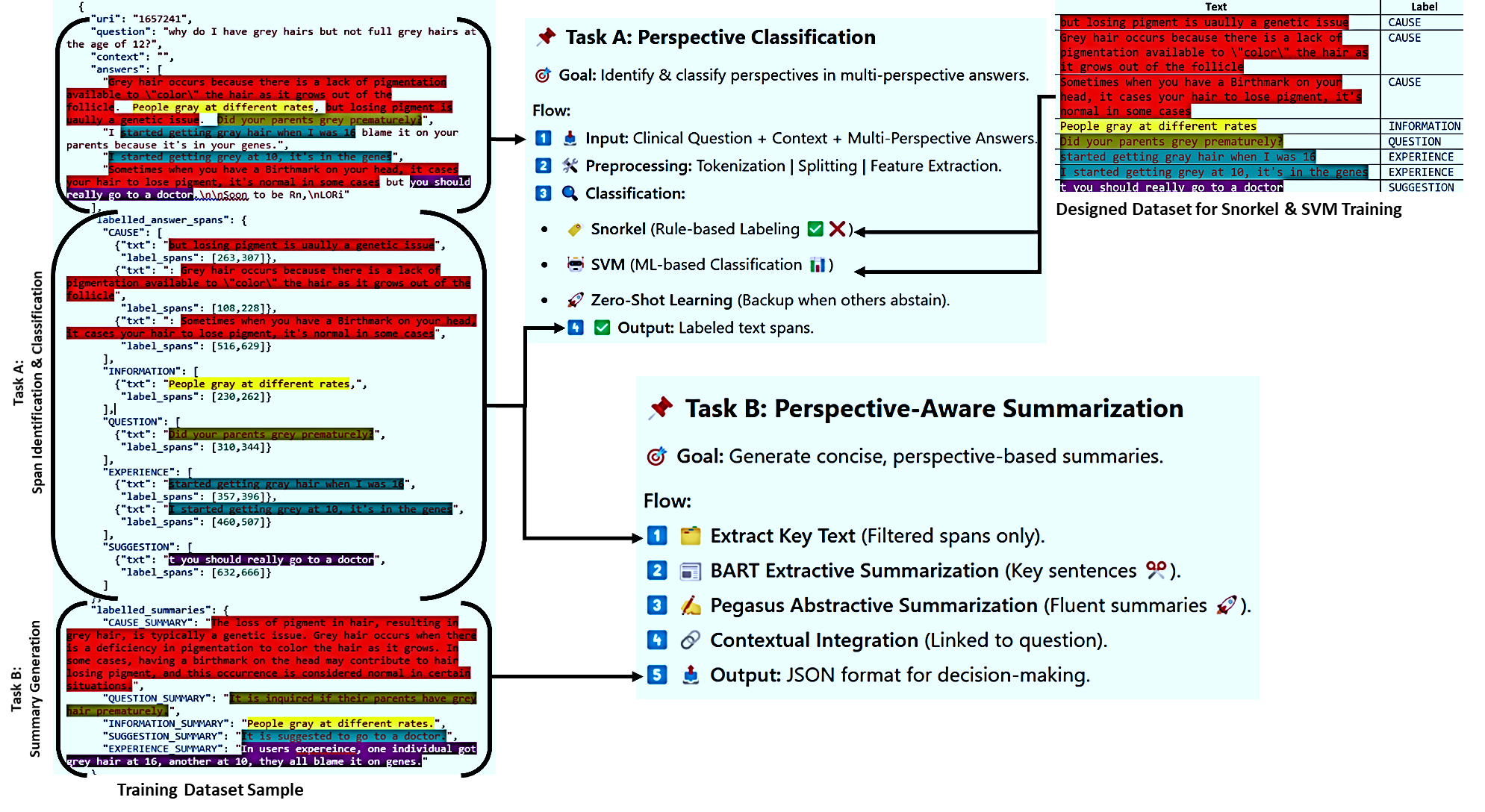}
  \caption{Training sample utilization for weak supervision. Known text spans from labeled data are used to train an SVM classifier, construct Snorkel labeling functions, and refine heuristic rules. The zero-shot model is excluded from direct training and is used as a fallback during classification.}
  \label{fig:03}
\end{figure*}
\subsubsection{Extractive Summarization Using BART}
We use the facebook/bart-large-cnn model to extract salient content \cite{lewis2019bart}:

\begin{equation}
 S = \text{BART}(X)
\end{equation}

where $X$ is the concatenated input text and $S$ is the generated extractive summary. The process involve following steps as shown in Figure \ref{fig:02} and \ref{fig:03}:

1. Tokenizing input text with BART’s tokenizer.

2. Using a task-specific prefix (summarize:).

3. Truncating text to 1024 tokens.

4. Applying beam search with: \textit{
\text{max\_length} = 150, \quad \text{min\_length} = 50, \quad \text{length\_penalty} = 2.0, \quad \text{num\_beams} = 4}

\subsubsection{Abstractive Refinement Using Pegasus}
The extractive summary is refined with google/pegasus-xsum \cite{zhang2020pegasus}:
\begin{equation}
 S' = \text{Pegasus}(S)
\end{equation}
where $S'$ is the final abstractive summary. Refinement involves following steps:

1. Tokenizing extractive summaries.

2. Using the summarize: prompt.

3. Truncating input to 512 tokens.

4. Applying beam search with:
        \textit{\text{max\_length} = 100, \quad \text{min\_length} = 30, \quad \text{length\_penalty} = 1.8, \quad \text{num\_beams} = 6}

For our experiments, we utilize a dataset labeled with five perspective categories \textit{P} in which \textit{EXPE} and all others relate to the perspective of Experience, Information, Cause, Suggestion, and Question respectively (in Equation \ref{eq.1}). Task A involves hierarchical classification, where unlabeled responses are processed using a combination of weak supervision, Support Vector Machines (SVM), and zero-shot learning (ZSL) (as Equation \ref{eq.8}). We employ Snorkel for weak supervision, training its label model for 500 epochs to aggregate multiple labeling sources. Sentence embeddings are generated using SentenceTransformer (\textit{all-MiniLM-L6-v2}) \cite{lewis2019bart}, which serves as input to an SVM classifier trained with a linear kernel and default hyperparameters. For ZSL, we use Facebook’s BART-Large-MNLI to directly infer category labels from textual descriptions.

Task B focuses on response structuring and refinement using transformer-based summarization models. We employ BART-Large-CNN for extractive summarization, generating concise representations of textual responses. To enhance coherence and fluency, we further refine these summaries using Pegasus-XSum \cite{zhang2020pegasus}, an abstractive summarization model designed for extreme summarization tasks. The dataset for Task B consists of both labeled and unlabeled responses, allowing the models to learn from structured examples while refining free-text inputs. Our approach integrates both extractive and abstractive summarization techniques to ensure a well-structured and contextually rich final output.

\section{Results and Discussion}

In this study, we evaluated multiple hybrid models integrating Few-shot learning, weak supervision (Snorkel), and transformer-based architectures (BART, PEGASUS, and SVMs) for Span Identification \& Classification (Task A) and Summarization (Task B). The primary objective was to assess the effectiveness of different learning paradigms in handling biomedical text processing challenges. 
 \begin{figure*}[h]
  \includegraphics[width=\textwidth]{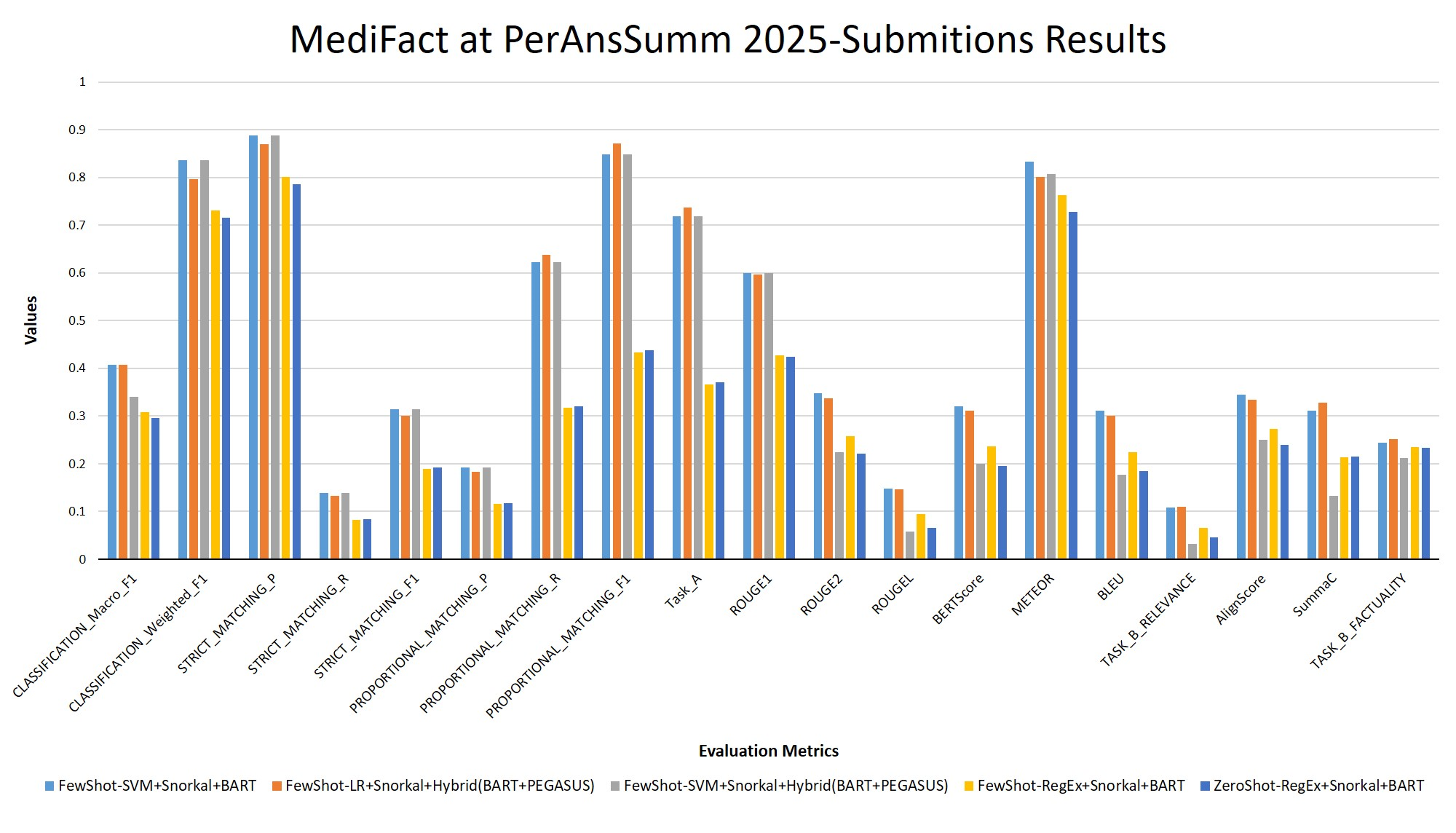}
  \caption{The comparative analysis of MediFact's submitted models on the PerAnsSumm Shared Task - CL4Health@NAACL 2025.  }
    \label{fig:nlp_detection}
\end{figure*}

Task A (Perspective Classification) is evaluated using Macro-F1, Weighted-F1, Strict Matching (Precision, Recall, Weighted-F1), and Proportional Matching (Precision, Recall, Weighted-F1). Task B (Perspective Summarization) is assessed using ROUGE (R1, R2, RL), BLEU, Meteor, and BERTScore. The bar graph illustrates a comparative analysis of model performance across both tasks, highlighting strengths and areas for improvement in Figure \ref{fig:nlp_detection}.

\subsection{Task A: Span Identification \& Classification}

The highest Weighted F1 score of 0.8361 was achieved by the \textit{FewShot-SVM+Snorkel+BART} model, demonstrating its robustness in span identification and classification. Additionally, \textit{FewShot-LR+Snorkel+Hybrid (BART+PEGASUS)} exhibited a competitive performance with an F1 score of 0.7961, while also achieving the best proportional match score (0.7373), indicating its capability to identify partially matched spans effectively.

Conversely, models relying on regular expressions (\textit{FewShot-RegEx+Snorkel+BART} and \textit{ZeroShot-RegEx+Snorkel+BART}) underperformed in classification, with \textit{F1} scores of 0.7316 and 0.7161, respectively. This suggests that rule-based approaches lack the generalization needed for complex biomedical text extraction tasks.

\subsection{Task B: Summarization Performance}

The summarization capabilities of the models were evaluated using ROUGE-1 scores and factuality assessments. The \textit{FewShot-SVM+Snorkel+BART} model achieved the highest ROUGE-1 score of 0.3485, indicating its effectiveness in generating relevant and concise summaries. Interestingly, \textit{FewShot-LR+Snorkel+Hybrid (BART+PEGASUS)} demonstrated superior factuality (0.2897), suggesting that PEGASUS contributes to improved content faithfulness in biomedical text summarization.

Models utilizing regular expression-based classification (FewShot-RegEx and ZeroShot-RegEx variants) performed significantly lower across all summarization metrics. This highlights that statistical and deep learning-based models outperform rule-based approaches in abstractive summarization tasks.

\subsection{Comparative Analysis of Model Performance}

For a comprehensive evaluation, the combined average score (Task A + Task B performance) was computed for each model (Figure \ref{fig:nlp_detection}). \textit{FewShot-SVM+Snorkel+BART} emerged as the best-performing approach with a combined score of 0.4077, followed by \textit{FewShot-LR+Snorkel+Hybrid (BART+PEGASUS)} with 0.4070. The hybrid models demonstrated a balanced trade-off between classification accuracy and summarization quality, reinforcing the effectiveness of weak supervision with Snorkel and transformer-based architectures.
In contrast, rule-based models (FewShot-RegEx \& ZeroShot-RegEx variants) consistently showed inferior performance, suggesting that deep generative models are more suitable for biomedical NLP tasks requiring contextual understanding and content generation.

The experimental results demonstrate that a hybrid learning strategy combining weak supervision (Snorkel), Few-shot learning, and transformer models (BART, PEGASUS) yields optimal performance in biomedical span identification and summarization tasks. The proposed \textit{FewShot-SVM+Snorkel+BART} model outperformed all other configurations, achieving the highest classification accuracy and summarization quality. These findings emphasize the importance of leveraging both structured supervision and deep generative architectures for enhancing biomedical text processing.

\subsection{MediFact Performance in PerAnsSumm Shared Task}
\label{sec:appendix}

MediFact secured a position among the \textbf{top 12 teams} in the \textbf{PerAnsSumm Shared Task - CL4Health@ NAACL 2025}. The final results were officially reported by the task organizers on the shared task website.\footnote{PerAnsSumm Shared Task - CL4Health@ NAACL 2025: \url{https://peranssumm.github.io/docs/\#leaderboard}}
 \begin{figure*}[ht]
  \includegraphics[width=\textwidth]{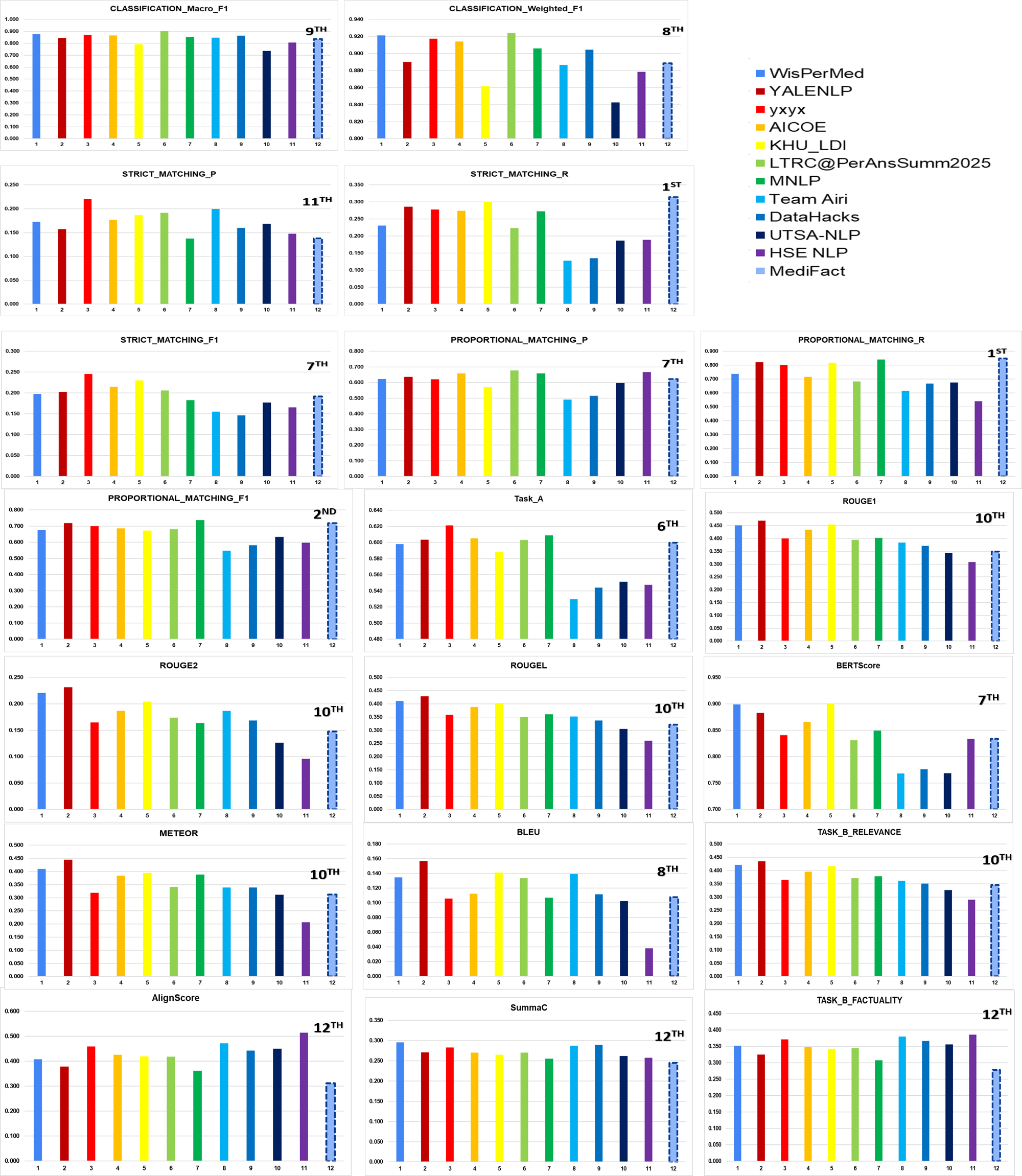}
  \caption{Comparative Performance Analysis of MediFact Among the Top 12 Models in the PerAnsSumm Shared Task – CL4Health@NAACL 2025.  }
    \label{fig:nlp_detection1}
\end{figure*}

In Figure \ref{fig:nlp_detection1}, MediFact's performance across various evaluation metrics demonstrates strong classification capabilities, achieving a competitive Weighted F1-score of 0.8887. However, the Macro F1-score (0.8361) suggests room for improvement in handling class imbalances. 

In the matching task, MediFact attains a high Proportional Matching Recall (0.8493), indicating effective identification of relevant matches. However, the Strict Matching Precision (0.1383) and Strict Matching F1 (0.1921) highlight challenges in reducing false positives. 

For summarization, the model achieves a BERTScore of 0.8336, reflecting strong semantic alignment. However, lower ROUGE scores (R1: 0.3485, R2: 0.1475, RL: 0.3212) and BLEU (0.1078) suggest the need for more accurate and concise text generation. 

Factual consistency metrics, such as AlignScore (0.3121) and Factuality Score (0.2784), indicate areas for improvement in ensuring reliable summarization. Future work should focus on enhancing precision in matching, optimizing summarization coherence, and strengthening factual alignment to ensure more trustworthy outputs.

\section{Conclusion}
This research introduces a modular and resource-efficient approach for perspective-aware classification and summarization. We combine weak supervision, machine learning, and pre-trained transformers to balance accuracy and computational cost \cite{ratner2017snorkel,rueping2010svm,lewis2019bart}. Instead of training a model from scratch, we fine-tune pre-trained models on our dataset. This approach reduces resource demands and speeds up adaptation to new tasks.  

One major motivation for our method is overcoming computational limitations. Training large models from the ground up requires extensive hardware and time \cite{touvron2023llama,floridi2020gpt,lewis2019bart}. To handle this, we use pre-trained models that can be fine-tuned efficiently. We also apply weak supervision with heuristic labeling, reducing the need for manual annotation (as shown in Figure \ref{fig:03}. This makes our approach scalable and practical.  

Our study shows that strong results can be achieved even with limited resources. We propose a modular and adaptable solution that does not depend entirely on commercial large language models (LLMs). While proprietary models offer high performance, they lack flexibility and accessibility \cite{team2023gemini,lee2020patent}. Instead, we demonstrate how open-source models and targeted fine-tuning provide robust results without heavy computational costs.  

There are some limitations to our approach. Weak supervision relies on heuristic rules, which may introduce bias or inconsistencies. While pre-trained models reduce the computational burden, further improvements can be made. Future research can explore lightweight architectures, efficient fine-tuning methods (such as LoRA \cite{hu2021lora} and quantization \cite{yang2019quantization}), and retrieval-augmented generation (RAG) \cite{notarangelo2016human} to handle unseen perspectives.  

In conclusion, this work highlights the importance of resource-aware AI research. It proves that effective NLP solutions can be built without expensive models. Open-source tools played a key role in making this study possible \cite{wolf2019huggingface,lewis2019bart,zhang2020pegasus}. By selecting the right model and designing a modular workflow, we achieve high-quality classification and summarization even with limited resources. This research encourages future work to focus on scalable, adaptable, and cost-effective AI solutions instead of relying solely on commercial LLMs.

\section*{List of Acronyms}

LR – Logistic Regression

SVM – Support Vector Machine

PEGASUS – Pre-trained Extractive Generative Abstractive SUmmarization System

BART - Bidirectional and Auto-Regressive Transformers

RegEx – Regular Expression

LoRA - Low-Rank Adaptation
\bibliography{custom}

\end{document}